\documentclass{article}

\usepackage[numbers]{natbib} 
\usepackage{makecell}

    \usepackage[preprint]{neurips_2025}

\usepackage{enumitem}

\usepackage[utf8]{inputenc} 
\usepackage[T1]{fontenc}    
\usepackage{hyperref}       
\usepackage{url}            
\usepackage{booktabs}       
\usepackage{amsfonts}       
\usepackage{nicefrac}       
\usepackage{microtype}      
\usepackage{xcolor}   

\usepackage{multirow}
\usepackage{wrapfig}
\usepackage{xspace}
\usepackage{graphicx}
\usepackage{amsmath}
\usepackage{mdframed}
\usepackage{tabularx}

\newmdenv[
  backgroundcolor=blue!3,
  linecolor=blue!50!black,
  linewidth=0pt,
  roundcorner=8pt,
  skipabove=12pt,
  skipbelow=12pt,
  innerleftmargin=12pt,
  innerrightmargin=12pt,
  innertopmargin=10pt,
  innerbottommargin=10pt,
  frametitle={\textsf{\textbf{PROMPT: Zero-Shot OOD Node Identification}}},
  frametitlefont=\sffamily\bfseries\small,
  frametitlerule=true,
  frametitlebackgroundcolor=blue!10,
  shadow=false
]{identpromptbox}

\newmdenv[
  backgroundcolor=green!2,
  linecolor=green!50!black,
  linewidth=0pt,
  roundcorner=8pt,
  skipabove=12pt,
  skipbelow=12pt,
  innerleftmargin=12pt,
  innerrightmargin=12pt,
  innertopmargin=10pt,
  innerbottommargin=10pt,
  frametitle={\textsf{\textbf{PROMPT: LLM-Based OOD Node Generation}}},
  frametitlefont=\sffamily\bfseries\small,
  frametitlerule=true,
  frametitlebackgroundcolor=green!10,
  shadow=false
]{genpromptbox}

\newcommand{\method}{{GOE-LLM}\xspace}
\newcommand{\methodid}{{GOE-identifier}\xspace}
\newcommand{\methodgen}{{GOE-generator}\xspace}

\setlist[itemize]{leftmargin=*, itemindent=0pt, itemsep=2pt, topsep=2pt}
\setlist[enumerate]{leftmargin=*, itemsep=2pt, topsep=2pt}
\setlist[description]{leftmargin=*, itemsep=2pt, topsep=2pt}

\title{Graph Synthetic Out-of-Distribution Exposure with Large Language Models}

\author{
\textbf{Haoyan Xu$^{1,*}$, 
Zhengtao Yao$^{1,*}$, 
Ziyi Wang$^2$,
Zhan Cheng$^3$,}\\
\textbf{Xiyang Hu$^4$, 
Mengyuan Li$^1$, 
Yue Zhao$^1$} \\
$^1$University of Southern California \quad
$^2$University of Maryland, College Park \quad \\
$^3$University of Wisconsin, Madison \quad
$^4$Arizona State University \\
\texttt{haoyanxu@usc.edu, zyao9248@usc.edu, zoewang@umd.edu, zcheng256@wisc.edu,} \\
\texttt{xiyanghu@asu.edu, mengyuanli@usc.edu, yzhao010@usc.edu}
}

\begin{document}

\maketitle

\begingroup
\renewcommand\thefootnote{*}
\footnotetext{Equal contribution.}
\endgroup

\begin{abstract}
Out-of-distribution (OOD) detection in graphs is critical for ensuring model robustness in open-world and safety-sensitive applications. 
Existing graph OOD detection approaches typically train an in-distribution (ID) classifier on ID data alone, then apply post-hoc scoring to detect OOD instances.
While \emph{OOD exposure}—adding auxiliary OOD samples during training—can improve detection, current graph-based methods often assume access to real OOD nodes, which is often impractical or costly.
In this paper, we present \method, a framework that leverages Large Language Models (LLMs) to achieve OOD exposure on text-attributed graphs without using any real OOD nodes.
\method introduces two pipelines: (1) identifying pseudo-OOD nodes from the initially unlabeled graph using zero-shot LLM annotations, and (2) generating semantically informative synthetic OOD nodes via LLM-prompted text generation.  
These pseudo-OOD nodes are then used to regularize ID classifier training and enhance OOD detection awareness. 
Empirical results on multiple benchmarks show that \method substantially outperforms state-of-the-art methods without OOD exposure, achieving up to a 23.5\% improvement in AUROC for OOD detection, and attains performance on par with those relying on real OOD labels for exposure.
\end{abstract}

\section{Introduction}
Graph data is widely used to model interactions among entities in social networks, citation networks, transaction networks, recommendation systems, and biological networks \cite{xiao2020timme, zhu2022survey, xu2021graph, xu2020cosimgnn, lee2020autoaudit}. In many practical scenarios, nodes are paired with rich textual attributes—such as user bios, paper abstracts, or product descriptions—leading to \textit{text-attributed graphs} (TAGs) \cite{yang2021graphformers, yan2023comprehensive}. These graphs integrate both structural and semantic information, enabling more fine-grained learning and inference tasks.
Recently, \emph{out-of-distribution (OOD) detection} on graphs \cite{wu2023energy, marevisiting, xu2025few, zhao2020uncertainty, xu2024lego, wang2025gold, xu2025few} has become increasingly studied for safety-critical and open-world applications. 
The goal is to identify nodes whose distribution significantly deviates from the in-distribution (ID) training classes. This task is particularly relevant in real-world applications where unseen or anomalous entities may appear during inference—such as emerging users in social platforms and new research domains in citation graphs.

\begin{wrapfigure}{L}{0.5\textwidth} 
    \centering
    \includegraphics[width=1\linewidth]{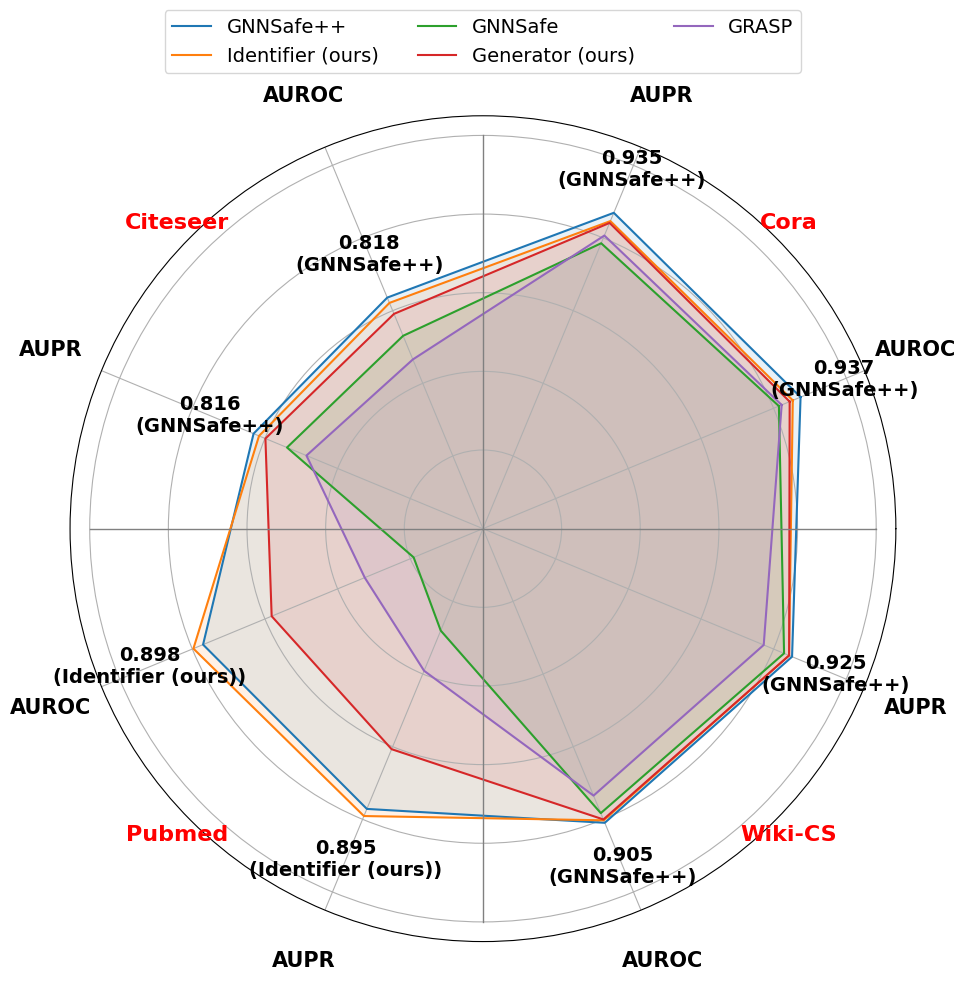}
    \caption{Our graph OOD detection method does not rely on any real OOD nodes for training, yet achieves significantly better OOD detection performance than baseline methods and performs comparably to the approach that uses real OOD nodes (GNNSafe++) for exposure.}
    \label{fig:Intro_img}
\end{wrapfigure}

\noindent
\textbf{OOD Exposure and Its Applicability to Graphs}.
Most existing graph OOD detection methods adopt a semi-supervised, transductive setup in which all nodes are accessible during training, but only a subset of classes is labeled \cite{marevisiting, song2022learning, xu2024lego}.
Training solely on ID data can lead to overconfident predictions on OOD nodes \cite{inkawhich2021training}, making subsequent post-hoc OOD scoring less reliable.
It can be even worse if OOD instances are structurally or semantically similar to ID data. 
A widely-used strategy for mitigating this overconfidence is \emph{OOD exposure}, wherein additional OOD samples are incorporated during training \cite{hendrycks2022pixmix,hendrycks2018deep,zhang2023mixture, du2024does}. 
However, existing approaches often rely on real OOD labels—an assumption that is unrealistic in many graph settings \cite{wu2023energy}, where OOD nodes are elusive or costly to label.
In other domains (images, text), recent work has investigated generating \emph{pseudo-OOD} instances \cite{tao2023non, abbas2025out, cao2024envisioning, du2022vos, wang2020further} to reduce reliance on real OOD data. 
For example, Large Language Models (LLMs) have been used to create OOD proxies for text detection \cite{abbas2025out} or outlier exposure \cite{cao2024envisioning}, yet these techniques are not directly applicable to graph data due to the complexity of node interconnections.

\noindent
\textbf{Our Proposal: \texttt{\methodid} and \texttt{\methodgen}.}
In this paper, we address OOD detection on TAGs by leveraging LLMs to generate OOD supervision signals without using real OOD samples. 
Specifically, we design two pipelines that inject pseudo-OOD information into training:
\underline{First: \texttt{\methodid}.} We randomly sample a small set of unlabeled nodes and prompt an LLM for zero-shot OOD detection. 
If the LLM concludes that a node does not match any known ID classes, it labels that node as “none,” effectively designating it OOD. 
Despite potential label noise, these \emph{identified} OOD nodes are then used as auxiliary training data to regularize the ID classifier.
\underline{Second: \texttt{\methodgen}.} Instead of annotating existing nodes, we instruct the LLM to \emph{generate} entirely new pseudo-OOD nodes. 
These are inserted into the original graph, along with corresponding OOD labels, to enhance the model’s ability to separate OOD from ID classes. 
Fig.~\ref{fig:Intro_img} shows that both strategies outperform baselines w/o OOD exposure, while matching approaches with real OOD data.

\textbf{\texttt{\methodid} vs. \texttt{\methodgen}.} 
In practice, \texttt{\methodid} is convenient when a graph already provides sufficient semantic context, allowing the LLM to reliably mark OOD nodes from unlabeled data. 
By contrast, \texttt{\methodgen} is preferable if node text is limited or if broad OOD concepts must be introduced. 
It also applies more naturally in inductive scenarios where future nodes—unseen at training time—may be OOD.
Meanwhile, \texttt{\methodid} targets a transductive setting, since the potential OOD nodes must exist beforehand. 
Overall, \texttt{\methodid} provides a lightweight way to detect OOD within the given graph, whereas \texttt{\methodgen} synthesizes new OOD samples when no suitable OOD data or domain knowledge is available in the unlabeled set.

We summarize our key contributions as follow:
\begin{itemize}[leftmargin=*, itemindent=0pt]
\item 
\textbf{First Method for LLM-Powered Graph OOD Exposure.}
We present a new approach to graph OOD detection with exposure that does not need real OOD labels, leveraging LLMs for pseudo-OOD identification and generation.

\item \textbf{Exploration of LLM Roles in OOD Exposure}. 
We propose two approaches for pseudo-OOD supervision: using an LLM as a pseudo-OOD node \textit{identifier} and as a pseudo-OOD node \textit{generator}. 
\texttt{\methodid} is effective in the transductive setting with rich unlabeled node semantics and requires no prior OOD knowledge. \texttt{\methodgen} is applicable to the inductive setting and can introduce novel OOD concepts, but it benefits from a global understanding of OOD semantics.
\item 
\textbf{Effectiveness}.
Experimental results demonstrate that our method significantly outperforms baselines without OOD exposure and performs similarly to methods that use real OOD nodes for exposure. 
\end{itemize}

\section{Related Work}
\subsection{Graph OOD Detection}
Detecting OOD samples 
has been extensively investigated in the graph domain. 
GNNSafe~\cite{wu2023energy} reveals that standard GNN classifiers inherently exhibit some ability to distinguish OOD nodes, and it proposes an energy-based discriminator trained with a standard classification objective. 
OODGAT~\cite{song2022learning} introduces a feature-propagation mechanism that explicitly separates inliers from outliers, unifying node classification and outlier detection in a single framework. 
GRASP~\cite{marevisiting} demonstrates the benefit of OOD score propagation and theoretical guarantees for post-hoc node-level OOD detection, supplemented by an edge-augmentation strategy. 
More recently, GNNSafe++~\cite{wu2023energy} extends GNNSafe by leveraging real OOD node labels for outlier exposure via an auxiliary regularization objective. 
However, methods that rely on actual OOD node labels are costly or infeasible, as identifying representative OOD nodes in graphs is non-trivial. 
Our work addresses the need for OOD exposure \emph{without} real OOD data, specifically on TAGs.


\subsection{OOD Detection with OOD Exposure}
A common strategy to mitigate overconfidence in neural networks is \emph{OOD exposure}, which incorporates OOD examples during training~\cite{hendrycks2018deep,yang2024generalized}. 
In image-based setups, several methods rely on real OOD samples or external datasets, sometimes using mixing strategies to expand the OOD coverage~\cite{hendrycks2022pixmix,zhang2023mixture}. 
For instance, OECC~\cite{papadopoulos2021outlier} appends a confidence-calibration term to further separate ID and OOD regions, and MixOE~\cite{zhang2023mixture} systematically mixes ID samples with known outliers to smooth the decision boundary. 
However, such techniques are limited by the availability and quality of genuine OOD data~\cite{du2024does}, which can be difficult or expensive to acquire in practice—particularly for graph data.

\textbf{Pseudo-OOD Generation.}
To circumvent the requirement for real OOD examples, recent efforts have explored \emph{pseudo-OOD generation}. 
VOS~\cite{du2022vos} synthesizes OOD representations from within the model’s latent space, and \cite{vernekar2019out} proposes a method to train an $(n+1)$-class classifier by generating OOD samples in the image domain. 
Likewise, EOE~\cite{cao2024envisioning} uses LLMs to envision outlier concepts for image-based OOD detection, while \cite{abbas2025out} uses LLMs to create high-quality textual proxies for text OOD detection. 
Despite their successes, these approaches typically overlook graph-specific challenges, where node connectivity and structural information must be modeled alongside semantic content. 
Indeed, image and text instances are mostly independent, but graph nodes have relational dependencies among neighbors, making it non-trivial to adapt existing pseudo-OOD methods to the graph domain. 
In this paper, we focus on TAGs and propose using LLMs to identify and generate OOD nodes without relying on any real OOD data,  bridging the gap left by prior work that has concentrated on images, text, or real OOD nodes in graphs.
A detailed comparison of the above methods
is presented in Appendix~\ref{appen:methods comparison}.

\section{Methodology}
\label{sec:methodology}

\begin{figure*}[!t]
   \begin{center}
   \includegraphics[width=1\linewidth]{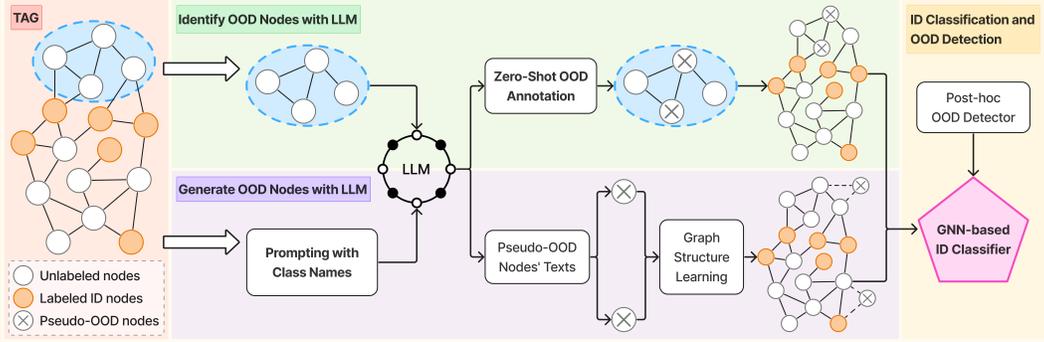} %
    \end{center}
    \vspace{-0.1in}
\caption{An overview of our framework \method. We design two methods for graph OOD exposure without using real OOD nodes: LLM as OOD nodes' identifier and LLM as OOD nodes' generator. The first method employs an LLM to identify potential OOD nodes from the initially fully unlabeled graph. These nodes are then used as pseudo-OOD samples to regularize the training of the ID classifier. The second method prompts the LLM to generate textual descriptions of synthetic OOD nodes, which are subsequently embedded and inserted into the original graph to enrich OOD information during ID classifier training.
}
\label{fig:framwork}
\end{figure*}

Here, we introduce \method, a framework that integrates pseudo-OOD exposure into TAG learning without requiring real OOD nodes. Fig.~\ref{fig:framwork} summarizes the overall pipeline.
Section~\ref{subsec:OOD exposure} describes the general process of graph OOD exposure, which uses real or pseudo-OOD nodes to regularize the training of the ID classifier.
Sections~\ref{subsec:goe-identifier} and~\ref{subsec:goe-generator} detail our proposed methods for graph OOD exposure without real OOD nodes: using an LLM as an OOD node identifier and as an OOD node generator, respectively.
Finally, section~\ref{sec:synthetic model} discusses additional strategies for training a synthetic OOD model to incorporate OOD information into the ID classifier.

\subsection{Preliminaries}
Our focus is on node-level OOD detection in TAGs. A TAG is represented as 
$
G_T = (\mathcal{V}, \mathbf{A}, \mathbf{T}, \mathbf{X}),
$
where \(\mathcal{V} = \{v_1, \dots, v_n\}\) is the set of \(n\) nodes, each paired with textual attributes \(\mathbf{T} = \{t_1, \dots, t_n\}\). We obtain sentence embeddings \(\mathbf{X} = \{x_1, \dots, x_n\}\) (e.g., via SentenceBERT~\cite{reimers2019sentence}), and \(\mathbf{A} \in \{0, 1\}^{n \times n}\) is the adjacency matrix, where \(\mathbf{A}[i, j] = 1\) indicates an edge between nodes \(v_i\) and \(v_j\).

\noindent \textbf{Node-level Graph OOD Detection}.
We split the nodes into a labeled set \(\mathcal{V}_I\) (ID nodes) and an unlabeled set \(\mathcal{V}_U\), with \(\mathcal{V}_I \cap \mathcal{V}_U = \emptyset\) and \(\mathcal{V}_I \cup \mathcal{V}_U = \mathcal{V}\). 
Each labeled node belongs to one of \(K\) known ID classes \(Y_I = \{y_1, \dots, y_K\}\). 
Unlabeled nodes may be either ID or OOD (i.e., from unknown classes not in \(Y_I\)). We adopt a semi-supervised, transductive setting: the entire graph is observable at training time, but only \(\mathcal{V}_I\) is labeled. 
The goal is to decide, for each \(v_i \in \mathcal{V}_U\), whether it belongs to one of the ID classes or is OOD.

\subsection{Graph OOD Detection with Pseudo-OOD Exposure}
\label{subsec:OOD exposure}
\noindent
\textbf{ID Classifier}.
OOD exposure incorporates OOD-like samples during training to tighten the boundary around ID data. 
We first train a two-layer GCN classifier using only ID-labeled nodes (\( \mathcal{V}_I \)):

\begin{equation}
    \mathcal{L}_{\text{sup}} \;=\; -\frac{1}{|\mathcal{V}_I|} \sum_{i \in \mathcal{V}_I} \sum_{k=1}^{K} y_{ik}\,\log \hat{y}_{ik},
    \label{equation::classifier}
\end{equation}
where \(\hat{y}_{ik}\) is the predicted probability that node \(v_i\) belongs to class \(k\).
After training, any post-hoc OOD scoring function can be applied. 
One common choice is the negative energy score~\cite{liu2020energy}, where the OOD score for a node \(v_i\) with logits \(z_i\in\mathbb{R}^K\) is:
\begin{equation}
    S_{\text{OOD}}(v_i) \;=\; -E(v_i) 
    \;=\; \log \sum_{k=1}^K \exp\bigl(z_{ik}\bigr).
    \label{eqn:OOD Score}
\end{equation}
A larger \(S_{\text{OOD}}(v_i)\) indicates that \(v_i\) is more likely OOD.

\noindent
\textbf{Pseudo-OOD Exposure.}
To refine the boundary between ID and OOD, we augment the training set with a set \(\mathcal{V}_O\) of OOD-like nodes. 
We then introduce a regularization term \(\mathcal{L}_{\text{expo}}\) that enforces contrasting OOD scores for ID nodes and OOD nodes. 
Specifically, we penalize ID nodes whose OOD scores exceed a margin \(s_{\text{id}}\) and OOD nodes whose OOD scores fall below a margin \(s_{\text{ood}}\). The overall loss is 
$
\mathcal{L}_{\text{sup}} + \lambda \,\mathcal{L}_{\text{expo}},
$
where \(\lambda\) balances ID classification accuracy and OOD discriminability. 
\begin{equation}
\label{equation:reg}
\mathcal{L}_\text{expo} = \frac{1}{|\mathcal{V}_I|} \sum_{v_i \in \mathcal{V}_I} \left( \operatorname{ReLU} \left( S_{\text{OOD}}(v_i) - s_{\text{id}} \right) \right)^2
+ \frac{1}{|\mathcal{V}_O|} \sum_{v_j \in \mathcal{V}_O} \left( \operatorname{ReLU} \left( s_{\text{ood}} - S_{\text{OOD}}(v_j) \right) \right)^2
\end{equation}
where \(s_{\text{id}}\) and \(s_{\text{ood}}\) are margin parameters, and \(\mathcal{V}_O\) can consist of either human-annotated OOD nodes or synthetic ones derived from LLMs.

\subsection{Identify OOD Nodes with LLMs}
\label{subsec:goe-identifier}
In this setting, rather than relying on real OOD nodes to train the ID classifier, we assume a more challenging and realistic scenario where there is no access to any OOD nodes or even the names of OOD classes. To address this, we leverage the transductive nature of graph learning and propose a novel approach that instructs LLMs to identify potential OOD nodes directly from the original graph. The identified pseudo-OOD nodes are then used to regularize the training of the ID classifier.
Specifically, we first randomly sample a small set of nodes $\mathcal{V}_U^{\text{sampled}}$ from $\mathcal{V}_U$ and let the LLM annotate them. The LLM is only provided with ID knowledge (ID class names) and prompted to determine whether the unlabeled query node belongs to an ID class, using its textual information.  
We instruct the LLM to output "none" if it determines that the node does not belong to any predefined ID class. After that, we select the nodes identified by the LLM as OOD to form the pseudo-OOD node set $\mathcal{V}_O$, as shown in Eqn.~\ref{equation:set1}. Using this annotated set $\mathcal{V}_O$, we then train the ID classifier with Eqn.~\ref{equation:reg}.
\begin{equation}
\mathcal{V}_O = \{ v \in \mathcal{V}_U^{\text{sampled}} \mid \text{LLM}(\text{prompt}(v, \mathcal{C}_{\text{id}})) = \text{"none"} \}
\label{equation:set1}
\end{equation}
where $\mathcal{C}_{\text{id}}$ denotes the set of ID category names.

\subsection{Generate OOD Nodes with LLMs}
\label{subsec:goe-generator}
Instead of using LLMs to identify potential OOD nodes in the graph, we propose an alternative approach that leverages LLMs to generate pseudo-OOD nodes and insert them into the original graph for OOD supervision. These generated nodes constitute the OOD instances \(\mathcal{V}_O\) as defined in Eqn.~\ref{equation:reg}. 
In this setting, we assume access only to the label names of all classes, without any real OOD nodes. For each OOD class, we use the LLM to generate \( M \) samples, leveraging its inherent large-scale knowledge of the respective class domains. The resulting text descriptions of the pseudo-OOD nodes are denoted by \(\mathbf{T}^{\text{ood}}\), and the generation process is formalized in Eqn.~\ref{eqn:OOD generation}.

\begin{equation}
\label{eqn:OOD generation}
    \mathcal{V}_O = \{ v_m^{\text{ood}} \mid \mathbf{T}_m^{\text{ood}} = \text{LLM}(\text{prompt}(c)),\ 1 \leq m \leq M,\ c \in \mathcal{C}_{\text{ood}} \}
\end{equation}
where $\mathcal{C}_{\text{ood}}$ denotes the set of OOD category names. 
By applying SentenceBERT \cite{reimers2019sentence} to \( \mathbf{T}^{\text{ood}} \), we obtain the embeddings of the pseudo-OOD nodes as \( \mathbf{X}^{\text{ood}} \). The complete set of node embeddings is then given by \( \mathbf{X}_{\text{aug}} = \mathbf{X} \mathbin{\|} \mathbf{X}^{\text{ood}} \).
Optionally, with graph structure learning, we can incorporate the pseudo-OOD nodes into the original graph to better propagate information through the new graph structure, denoted as \( \mathbf{A}_{\text{aug}} \). One way to construct \( \mathbf{A}_{\text{aug}} \) is by creating edges based on the similarity of node embeddings. Alternatively, a negative sampling-based link prediction task can be performed to infer potential links. Using \( \mathbf{X}_{\text{aug}} \) and \( \mathbf{A}_{\text{aug}} \), we then train the ID classifier on the augmented graph. This approach reframes OOD detection as an active and generative strategy—rather than a passive inference task—by leveraging textual priors to construct meaningful semantic contrast without requiring real OOD data.
\vspace{-0.1in}

\subsection{Synthetic OOD Model}
\label{sec:synthetic model}
Thus far, we have proposed using pseudo-OOD nodes to regularize the training of a \( K \)-class ID classifier and applying post-hoc OOD detectors on top of the well-trained ID classifier for OOD detection. The main advantage of this approach is that it does not require modifying the network architecture of the ID classifier. However, it introduces a trade-off weight \(\lambda\) in the loss function. Intuitively, if \(\lambda\) is too large, it may degrade ID classification performance. Conversely, if \(\lambda\) is too small, the OOD information may not be sufficiently exposed to the ID classifier to improve its OOD awareness. To address this, we provide two alternative approaches that leverage both labeled ID nodes and synthetic noisy OOD nodes to train a model with enhanced OOD awareness. 

The first approach involves adding a binary classification layer on top of the standard ID classifier trained using Eqn.~\ref{equation::classifier} to predict OOD scores. Specifically, we first train the ID classifier using labeled ID nodes. Once trained, we freeze the ID classifier and fit the weights of the binary OOD detector using a small set of labeled ID nodes along with the pseudo-OOD samples. The key advantage of this method is that it preserves the ID predictions of the pre-trained classifier while equipping the model with the capability to detect OOD nodes through the additional binary layer. The second approach extends the ID classifier into a $(K + 1)$-way classification model, where the first $K$ classes correspond to the ID classes and the $(K + 1)$-th class represents the OOD class. The model is trained using both labeled ID nodes and pseudo-OOD nodes under a unified cross-entropy loss. The primary advantage of this joint classification approach lies in its flexibility to simultaneously learn accurate ID predictions while distinguishing between ID and OOD nodes, thereby enhancing overall performance.
\vspace{-0.2in}

\section{Experiments}
\label{sec:experiments}
\vspace{-0.1in}
\subsection{Experimental Setup}
\label{subsec:exp setup}
\noindent \textbf{Datasets}
We utilize the following TAG datasets, which are commonly used for node classification: Cora \cite{mccallum2000automating}, Citeseer \cite{giles1998citeseer}, Pubmed \cite{sen2008collective} and Wiki-CS \cite{mernyei2020wiki}. The dataset descriptions are in Appendix \ref{appen:dataset-desc}.
We follow previous work \cite{song2022learning} by splitting the node classes into ID and OOD classes, ensuring that the number of ID classes is at least two to support the ID classification task. The specific class splits and ID ratios are detailed in Appendix \ref{appen:OOD Split}. 

\noindent \textbf{Training and Evaluation Splits} For each dataset with $K$ ID classes, we use $20\times K$ of ID nodes for training. Additionally, we randomly select $10\times K$ of ID nodes along with an equal number of OOD nodes for validation. The test set consists of 500 randomly selected ID nodes and 500 OOD nodes. All experiments are repeated with five random seeds, and results are averaged.

\noindent \textbf{Baselines}
We compare \method with the following baselines: (1) \textbf{MSP} \cite{hendrycks2016baseline}, (2) \textbf{Entropy}, (3) \textbf{Energy} \cite{liu2020energy}, (4) \textbf{GNNSafe} \cite{wu2023energy}, and (5) \textbf{GRASP} \cite{marevisiting}, all of which are post-hoc OOD detection methods without exposure. Additionally, we include \textbf{OE} \cite{hendrycks2018deep} and \textbf{GNNSafe++} \cite{wu2023energy}, which leverage real OOD samples for exposure.

\noindent \textbf{LLM-Powered OOD Exposure.} We use GPT-4o-mini for both pseudo-OOD identification and generation. For node identification, we randomly sample 200 unlabeled nodes per dataset and prompt the LLM to classify them as either ID or OOD. Nodes predicted as OOD are used as pseudo-OOD exposure data. For pseudo-OOD generation, we prompt the LLM to generate $10 \times K_{\text{ood}}$ nodes, where $K_{\text{ood}}$ denotes the number of OOD classes.
The generated node texts are embedded using SentenceBERT \cite{reimers2019sentence}, and the resulting embeddings are concatenated with the original graph embeddings. We leave the exploration of more advanced graph structure learning methods for constructing the augmented graph to future work.

\noindent \textbf{Implementation details} For fair comparison, all ID classifiers are implemented using 2-layer GCNs with hidden dimension 32. We use Adam optimizer with learning rate 0.01, dropout rate 0.5, and  weight decay of 5\text{e-}4. For all OOD exposure methods, the trade-off weight $\lambda$ in the loss function is selected from $\{0.01,0.05\}$ based on the results of the validation set. For all methods, we set the maximum number of training epochs to 200 and apply early stopping if the sum of AUROC and ID ACC does not improve for 20 epochs. All experiments are conducted on hardware equipped with an NVIDIA GeForce RTX 4080 SUPER GPU.

\noindent \textbf{Evaluation Metrics}
For the ID classification task, we use classification accuracy (ID ACC) as the evaluation metric. For the OOD detection task, we employ three commonly used metrics from the OOD detection literature \cite{wu2023energy}: the area under the ROC curve (AUROC), the precision-recall curve (AUPR), and the false positive rate when the true positive rate reaches 95\% (FPR@95). In all experiments, OOD nodes are considered positive cases. Detailed descriptions of these metrics are provided in Appendix~\ref{appen:metrics}.


\subsection{Main Results}
\label{sec:main_results}

Table \ref{tab:results} presents the performance of various OOD detection methods across four datasets. From the results, we make several key observations:

\noindent \textbf{Effectiveness of LLM-driven OOD Exposure.}  
Both variants of \method—\methodid and \methodgen—consistently outperform all methods that do not utilize OOD exposure. For example, on the Pubmed dataset, \methodid achieves an AUROC of 0.8985, significantly surpassing GRASP (0.6627), the best-performing method without OOD exposure, resulting in a relative improvement of over 23.5\%. This demonstrates that the pseudo-OOD nodes identified by the LLM provide meaningful supervision for learning precise decision boundaries in open-world settings. However, the improvement on the Wiki-CS dataset is much smaller, suggesting that the effectiveness of pseudo-OOD exposure depends on the dataset's inherent difficulty and the extent of distributional shift. Due to space constraints, the standard deviation results are provided in Appendix \ref{appen::STD}.

\noindent \textbf{Comparable Performance to Real OOD Supervision.}  
Remarkably, \method achieves performance comparable to OE and GNNSafe++, both of which use real OOD nodes annotated by humans. On the Pubmed dataset, \method even surpasses OE and GNNSafe++ in terms of AUROC and AUPR, despite relying solely on noisy pseudo-OOD nodes. This demonstrates that LLMs can serve as a practical alternative to costly OOD data curation. We attribute this strong performance to the LLM's ability to synthesize semantically coherent yet distributionally distinct samples that effectively emulate the characteristics of true OOD data.

\noindent \textbf{ID Classification is Maintained.}  
A common concern with OOD exposure is its potential to degrade ID classification performance due to over-regularization. However, our results show that \method maintains strong ID classification accuracy across all datasets. For example, on Citeseer, \methodgen reaches 0.8552 ID accuracy, outperforming all other baselines. This indicates that pseudo-OOD nodes do not dilute the model’s ability to learn discriminative features for the ID task.

\noindent \textbf{Insights on LLM Annotations.}  
Although the OOD nodes identified by LLMs are relatively noisy (as shown in Section~\ref{sec:zero-shot OOD}), using these noisy nodes to regularize the training of the ID classifier still yields results on par with those obtained using real OOD nodes. 
The key intuition is that, while LLM-based annotations may be noisy, they are not arbitrary. In fact, they reflect a distributionally-aware semantic prior: nodes misclassified as OOD by the LLM often lie near the ID-OOD boundary and can serve as hard negatives. This is evident in our results, where even imprecise OOD exposure improves downstream detection performance significantly. This supports the hypothesis that OOD exposure does not need to be perfect to be effective—it merely needs to be informative enough to delineate boundaries in the feature space.

\noindent \textbf{Overall Impact.}  
Taken together, our findings suggest that LLM-driven pseudo-OOD exposure is a promising and scalable direction for graph OOD detection. It enables label-free OOD supervision, maintains strong ID classification, and yields competitive or superior results compared to both non-exposure and real-exposure methods.

\begin{table*}[t]
    \centering
    \caption{Performance comparison (best in \textbf{bold}, second-best in \underline{underline}) of different methods on ID classification and OOD detection tasks. \method achieves comparable performance to methods using real OOD nodes, while requiring no real OOD data and significantly outperforming methods without OOD exposure.
    }
    \vspace{0.1in}
    \label{tab:results}
    \renewcommand{\arraystretch}{1.6} 
    \setlength{\tabcolsep}{1pt}      
    \resizebox{\textwidth}{!}{
    \begin{tabular}{llcccccccccccccccc}
        \toprule
        \multirow{2}{*}{\textbf{Model}} & \multirow{2}{*}{\textbf{Methods}} 
            & \multicolumn{4}{c}{\textbf{Cora}} 
            & \multicolumn{4}{c}{\textbf{Citeseer}} 
            & \multicolumn{4}{c}{\textbf{Pubmed}} 
            & \multicolumn{4}{c}{\textbf{Wiki-CS}} \\
        \cmidrule(lr){3-6}\cmidrule(lr){7-10}\cmidrule(lr){11-14}\cmidrule(lr){15-18}
        & & ID ACC $\uparrow$ & AUROC $\uparrow$ & AUPR $\uparrow$ & FPR95 $\downarrow$ 
          & ID ACC $\uparrow$ & AUROC $\uparrow$ & AUPR $\uparrow$ & FPR95 $\downarrow$
          & ID ACC $\uparrow$ & AUROC $\uparrow$ & AUPR $\uparrow$ & FPR95 $\downarrow$
          & ID ACC $\uparrow$ & AUROC $\uparrow$ & AUPR $\uparrow$ & FPR95 $\downarrow$ \\
        \midrule
        \multirow{5}{*}{\shortstack{No OOD\\Exposure}} 
        & MSP 
            & 0.8748 & 0.8414 & 0.8506 & 0.6428
            & 0.8480 & 0.7466 & 0.7500 & 0.7976
            & 0.8776 & 0.6591 & 0.6623 & 0.8908
            & 0.8648 & 0.7772 & 0.7851 & 0.7440 \\
        & Entropy 
            & \underline{0.8800} & 0.8471 & 0.8549 & 0.6300
            & 0.8480 & 0.7655 & 0.7603 & 0.7244
            & 0.8776 & 0.6591 & 0.6623 & 0.8908
            & 0.8640 & 0.7823 & 0.7891 & 0.7440 \\
        & Energy 
            & 0.8788 & 0.8580 & 0.8648 & 0.5928
            & \underline{0.8504} & 0.7757 & 0.7754 & 0.7256
            & 0.8876 & 0.5861 & 0.5919 & 0.9296
            & 0.8648 & 0.7983 & 0.8056 & 0.7320 \\
        & GNNSafe 
            & \underline{0.8800} & 0.9073 & 0.9073 & 0.4084
            & \underline{0.8504} & 0.7654 & 0.7697 & 0.8004
            & \textbf{0.8916} & 0.5954 & 0.6403 & 0.8928
            & 0.8752 & 0.8915 & 0.9144 & 0.7928 \\
        & GRASP 
            & \underline{0.8800} & 0.9111 & 0.9034 & 0.3820
            & 0.8460 & 0.7329 & 0.7429 & 0.8560
            & \underline{0.8908} & 0.6627 & 0.6956 & 0.8780
            & 0.8768 & 0.8674 & 0.8864 & 0.7860 \\
        \midrule
        \multirow{2}{*}{Ours}
        & \methodid 
            & 0.8780 & 0.9264 & 0.9233 & 0.3268
            & 0.8444 & 0.8107 & 0.8082 & 0.7104
            & 0.8636 & \textbf{0.8985} & \textbf{0.8955} & \textbf{0.5108}
            & 0.8764 & 0.9014 & 0.9212 & 0.7728 \\
        & \methodgen 
            & 0.8792 & 0.9221 & 0.9208 & 0.3372
            & \textbf{0.8552} & 0.7956 & 0.7994 & 0.7328
            & 0.8820 & 0.7908 & 0.8035 & 0.7780
            & 0.8752 & 0.9002 & \underline{0.9214} & \underline{0.7288} \\
        \midrule
        \multirow{2}{*}{\shortstack{Real OOD\\Exposure}} 
        & OE & \textbf{0.8820} & \underline{0.9362} & \underline{0.9392} & \underline{0.3196} & 0.8384 & \textbf{0.8348} & \textbf{0.8254} & \textbf{0.6100} & 0.8512 & 0.8236 & 0.8314 & 0.7236 & \textbf{0.8832} & \textbf{0.9166} & 0.9171 & \textbf{0.4248} \\
        & GNNSafe++ 
            & 0.8792 & \textbf{0.9371} & \textbf{0.9347} & \textbf{0.2944}
            & 0.8464 & \underline{0.8180} & \underline{0.8157} & \underline{0.6832}
            & 0.8760 & \underline{0.8852} & \underline{0.8857} & \underline{0.5568}
            & \underline{0.8816} & \underline{0.9048} & \textbf{0.9254} & 0.7668 \\
        \bottomrule
    \end{tabular}
    }
\vspace{-0.1in}
\end{table*}

\subsection{Is the Current Graph OOD Detection Setting Practical?}
Since there is no established graph-specific OOD detection benchmark that models real distributional shifts among nodes, most existing work on node-level graph OOD detection simply assumes that nodes from randomly selected classes are designated as OOD. However, this setting has notable limitations. For example, if the ID classes are more heterogeneous or closely resemble the OOD classes, the ID classifier may fail to learn a well-defined decision boundary, resulting in high softmax confidence scores for OOD nodes.  
Additionally, if the set of ID classes spans a broad and diverse range of features, the model may naturally assign high confidence to OOD nodes, violating the common assumption that OOD nodes should receive lower confidence scores.
Therefore, using real or pseudo-OOD nodes can provide a more realistic and effective training signal for OOD detection. By explicitly exposing the model to samples that are semantically or structurally different from the ID distribution, we can help the classifier better distinguish between ID and OOD nodes. This leads to more calibrated confidence estimates and improved robustness under open-world scenarios.

\begin{figure*}[h]
   \begin{center}
   \includegraphics[width=1\linewidth]{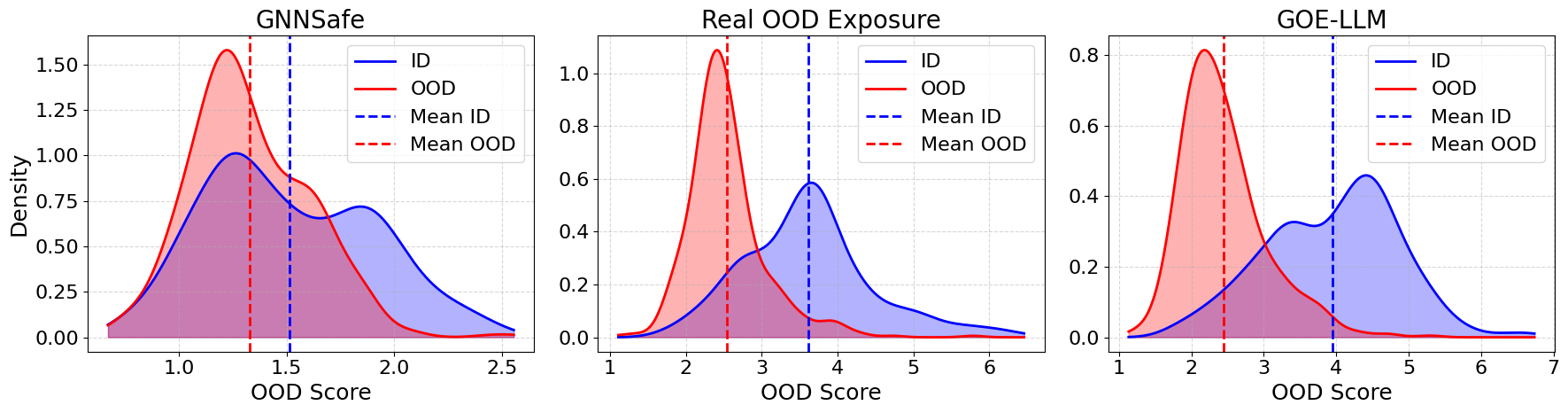} %
    \end{center}
    \vspace{-0.08in}
 \caption{OOD score distributions of GNNSafe, GNNSafe++, and \method on the Pubmed dataset. \method, despite not using any real OOD nodes, achieves distributional separation comparable to the real OOD exposure method and significantly better separability than GNNSafe.
}
\label{fig:OOD score}
\vspace{-0.05in}
\end{figure*}

To demonstrate this, we visualize the OOD scores on the Pubmed dataset for GNNSafe, \method, and GNNSafe++. As shown in Fig.~\ref{fig:OOD score}, compared to the method without OOD exposure, our approach assigns notably lower OOD scores to ID nodes and higher scores to OOD nodes, resulting in a more distinct separation between the two groups.

\subsection{Zero-shot OOD Annotation}
\label{sec:zero-shot OOD}
We present the zero-shot OOD detection performance of the LLM. The prompt used for zero-shot OOD node identification is provided in Prompt~\ref{identification_prompt}. For each dataset, we prompt the LLM to determine whether each node in the test set (comprising 500 ID nodes and 500 OOD nodes) belongs to one of the predefined ID categories; if not, the node is considered an OOD instance. We then obtain binary OOD predictions from the LLM and compute its accuracy. The results are reported in Table~\ref{tab:zero-shot}.
We use accuracy as the sole evaluation metric for the LLM's zero-shot OOD annotation performance, since the LLM does not produce soft OOD scores (e.g., as defined in Eqn.~\ref{eqn:OOD Score}), but rather outputs hard binary decisions (0 or 1).

\vspace{-0.05in}
\begin{identpromptbox}
\label{identification_prompt}
\ttfamily\footnotesize
\vspace{-0.05in}
As a research scientist, your task is to analyze and classify \{object\} based on their main topics, meanings, background, and methods. 

Please first read the content of the \{object\} carefully. Then, identify the \{object\}'s key focus. Finally, match the content to one of the given categories:

\texttt{[Category 1, Category 2, Category 3, ...]}

Given the current possible categories, determine if it belongs to one of them. If so, specify that category; otherwise, say "none".

[Insert \{Object\} Content Here]
\vspace{-0.05in}
\end{identpromptbox}

\vspace{-0.2in}

\begin{table}[h]

\centering
\begin{tabular}{lcccc}
\toprule
\textbf{} & \textbf{Cora}& \textbf{Citeseer} & \textbf{Pubmed} & \textbf{Wiki-CS}\\
\midrule
Zero-shot OOD annotation  & 0.7190 & 0.7260 & 0.8410 & 0.7470 \\
\bottomrule
\end{tabular}
\caption{Accuracy of using the LLM to identify whether unlabeled nodes are OOD. Although the LLM-identified pseudo-OOD nodes are noisy, they still enable effective OOD exposure.}
\label{tab:zero-shot}
\vspace{-0.2in}
\end{table}

\subsection{Synthetic Data Model}
In this section, we combine labeled ID nodes and pseudo-OOD nodes to train the synthetic data models, as described in Section \ref{sec:synthetic model}. Specifically, we randomly select $20 \times K$ ID nodes and pseudo-OOD nodes annotated by the LLM. In this way, we use the same number of ID and OOD nodes to train the synthetic data models.
\begin{table*}[h]
\vspace{-0.2in}
    \centering
    \caption{Performance comparison (best highlighted in bold) of different synthetic data models on ID classification and OOD detection tasks.}
    \label{tab:synthetic}
    \renewcommand{\arraystretch}{1.5}
    \setlength{\tabcolsep}{1pt}
    
    \resizebox{\textwidth}{!}{
    \begin{tabular}{lcccccccccccccccc}
        \toprule
        \multirow{2}{*}{\textbf{Methods}} 
        & \multicolumn{4}{c}{\textbf{Cora}} 
        & \multicolumn{4}{c}{\textbf{Citeseer}} 
        & \multicolumn{4}{c}{\textbf{Pubmed}} 
        & \multicolumn{4}{c}{\textbf{Wiki-CS}} \\
        \cmidrule(lr){2-5} \cmidrule(lr){6-9} \cmidrule(lr){10-13} \cmidrule(lr){14-17}
        & ACC $\uparrow$ & AUROC $\uparrow$ & AUPR $\uparrow$ & FPR95 $\downarrow$ 
        & ACC $\uparrow$ & AUROC $\uparrow$ & AUPR $\uparrow$ & FPR95 $\downarrow$
        & ACC $\uparrow$ & AUROC $\uparrow$ & AUPR $\uparrow$ & FPR95 $\downarrow$
        & ACC $\uparrow$ & AUROC $\uparrow$ & AUPR $\uparrow$ & FPR95 $\downarrow$ \\

        \midrule
        \method 
        & 0.8780 & 0.9264 & 0.9233 & 0.3268 
        & 0.8444 & 0.8107 & 0.8082 & 0.7104 
        & 0.8636 & 0.8985 & 0.8955 & 0.5108 
        & 0.8764 & 0.9014 & 0.9212 & 0.7728 \\
        
        
        $(K + 1)$-Classifier  
        & 0.8716 & 0.9138 & 0.9185 & 0.4224 
        & 0.8336 & 0.8189 & 0.8200 & 0.6648 
        & 0.8840 & 0.9060 & 0.8968 & 0.3980 
        & 0.8684 & 0.8015 & 0.7923 & 0.6648 \\
        
        \bottomrule
    \end{tabular}
    }
\vspace{-0.2in}
\end{table*}

For the first approach, we can add a binary classification layer on top of the output features of the ID classifier to predict the OOD score \( z_{\text{ood}} = \mathbf{w}^\top \phi(x) \), where \( \mathbf{w} \in \mathbb{R}^h \), and \( x \) is the output of the hidden layer from the GNN-based ID classifier. The OOD score of node \( v_i \) is then defined as \( S_{\text{OOD}}(v_i) = \sigma(z_{\text{ood}}) \), where \( \sigma(\cdot) \) denotes the sigmoid function.
In the second approach, we train a \( (K+1) \)-class classifier and define the softmax probability of the \( (K+1) \)-th class as the OOD score. However, the performance of the first approach is not satisfactory in the current setting; therefore, we only report the results of the \( (K+1) \)-class classifier.
The results are presented in Table \ref{tab:synthetic}. From the results, we observe that using Eqn.~\ref{equation:reg} to regularize the training of the ID classifier does not degrade ID classification performance. At the same time, it significantly improves OOD detection performance compared to the baselines without OOD exposure. Furthermore, the \( (K+1) \)-class classifier generally achieves performance comparable to that of the regularization method.

\subsection{How Many Synthetic OOD Nodes Do We Need?}
In this section, we prompt the LLM to generate different numbers of pseudo-OOD nodes. The prompt used for OOD node generation is provided in Prompt \ref{generation_prompt}.
Table \ref{tab:ood_exposure_comparison} presents the ID classification and OOD detection performance on the Pubmed dataset (which contains 19,717 nodes) using different numbers of generated pseudo-OOD nodes.
From the results, we can see that using more pseudo-OOD nodes improves OOD detection performance. When the number of generated OOD nodes reaches around 10, the performance nearly converges to a value that is significantly higher than that achieved without OOD exposure.
This suggests that even a small number of pseudo-OOD nodes can provide meaningful OOD exposure during training, helping the model learn a more accurate decision boundary between ID and OOD nodes. Moreover, it shows that LLM-generated pseudo-OOD nodes offer an efficient and lightweight substitute for real OOD data. Another observation is that, among all cases, the ID classification accuracy is highest when no OOD exposure is performed. However, the degradation in ID classification performance is negligible when pseudo-OOD nodes are used to regularize the training of the ID classifier. This further demonstrates that the trade-off described in Section \ref{subsec:OOD exposure} is favorable, as pseudo-OOD exposure significantly improves OOD detection performance while having minimal impact on ID classification accuracy.

\vspace{-0.1in}
\begin{table}[t]
\renewcommand{\arraystretch}{0.8}
\centering
\begin{tabular}{ccccc}
\toprule
\textbf{\makecell{Number of\\Pseudo-OOD Nodes}} & \textbf{ID ACC} $\uparrow$ & \textbf{AUROC} $\uparrow$ & \textbf{AUPR} $\uparrow$ & \textbf{FPR@95} $\downarrow$ \\
\midrule
\phantom{0} 0  & \textbf{0.8916} & 0.5954 & 0.6403 & 0.8928  \\
\phantom{0} 2  & 0.8864 & 0.6692 & 0.6961 & 0.8708 \\
\phantom{0} 3  & 0.8896 & 0.7401 & 0.7566 & 0.8100 \\
\phantom{0} 5  & 0.8796 & 0.7674 & 0.7857 & 0.8036 \\
\phantom{0} 10 & 0.8820 & 0.7908 & 0.8035 & 0.7780 \\
\phantom{0} 20 & 0.8856 & \textbf{0.8043} & \textbf{0.8073} & \textbf{0.7400} \\
\bottomrule
\end{tabular}

\caption{Performance comparison using different numbers of generated pseudo-OOD nodes for OOD exposure on the Pubmed dataset. Even a small number of pseudo-OOD nodes significantly improves OOD detection performance.}
\label{tab:ood_exposure_comparison}
\vspace{-0.25in}
\end{table}

\begin{genpromptbox}
\label{generation_prompt}
\ttfamily\footnotesize
Please generate \{num\_generated\_samples\} \{object\}(s) belonging to the category '\{category\_name\}', including title and abstract.

\textbf{Output Format:}
\begin{itemize}
    \item \texttt{Title: <Generated Title>}
    \item \texttt{Abstract: <Generated Abstract>}
\end{itemize}
\end{genpromptbox}
\vspace{-0.2in}

\section{Conclusion}
\vspace{-0.05in}
\label{sec:conclusion}
In this paper, we propose \method, a novel framework for OOD detection on TAGs that eliminates the reliance on real OOD data for exposure by leveraging LLMs. By designing two exposure strategies—LLM-based OOD node identification and OOD node generation—we enable label-efficient, scalable, and effective OOD exposure. Extensive experiments across diverse datasets demonstrate that \method achieves strong performance compared to methods relying on real OOD data, and significantly surpasses existing methods without OOD exposure. 
Future work could explore more advanced prompting strategies to improve the quality of pseudo-OOD samples. Additionally, incorporating adaptive graph structure learning tailored to generated nodes may further boost performance. Extending \method to other graph learning tasks—such as graph-level OOD detection and OOD detection on dynamic TAGs—also presents a promising direction.

\noindent \textbf{Broader Impact}.
On the positive side, this work provides a practical and generalizable solution to improve model reliability and robustness in open-world scenarios where labeled data is scarce or dynamic changes are frequent. This has the potential to enhance safety and trustworthiness in real-world deployments.
On the other hand, our reliance on LLM-generated supervision introduces challenges around semantic validity and bias. The quality and fairness of the pseudo-OOD labels depend on the pretrained LLMs, which may inherit societal or domain-specific biases.

\noindent \textbf{Limitations}. First, the effectiveness of pseudo-OOD nodes relies on the LLM’s zero-shot accuracy; inaccurate outputs may introduce noisy supervision. Second, while the proposed method markedly improves OOD detection performance on many graph datasets, it is applicable only to TAGs. For graphs without textual information, the method may be ineffective, highlighting the need for more generalizable OOD exposure techniques.

\bibliographystyle{plain}
\bibliography{references}

\begin{thebibliography}{10}

\bibitem{abbas2025out}
Momin Abbas, Muneeza Azmat, Raya Horesh, and Mikhail Yurochkin.
\newblock Out-of-distribution detection using synthetic data generation.
\newblock {\em arXiv preprint arXiv:2502.03323}, 2025.

\bibitem{cao2024envisioning}
Chentao Cao, Zhun Zhong, Zhanke Zhou, Yang Liu, Tongliang Liu, and Bo~Han.
\newblock Envisioning outlier exposure by large language models for out-of-distribution detection.
\newblock {\em arXiv preprint arXiv:2406.00806}, 2024.

\bibitem{dong2024multiood}
Hao Dong, Yue Zhao, Eleni Chatzi, and Olga Fink.
\newblock Multiood: Scaling out-of-distribution detection for multiple modalities.
\newblock {\em Advances in Neural Information Processing Systems}, 37, 2024.

\bibitem{du2024does}
Xuefeng Du, Zhen Fang, Ilias Diakonikolas, and Yixuan Li.
\newblock How does unlabeled data provably help out-of-distribution detection?
\newblock {\em arXiv preprint arXiv:2402.03502}, 2024.

\bibitem{du2022vos}
Xuefeng Du, Zhaoning Wang, Mu~Cai, and Yixuan Li.
\newblock Vos: Learning what you don't know by virtual outlier synthesis.
\newblock {\em arXiv preprint arXiv:2202.01197}, 2022.

\bibitem{giles1998citeseer}
C~Lee Giles, Kurt~D Bollacker, and Steve Lawrence.
\newblock Citeseer: An automatic citation indexing system.
\newblock In {\em Proceedings of the third ACM conference on Digital libraries}, pages 89--98, 1998.

\bibitem{hendrycks2016baseline}
Dan Hendrycks and Kevin Gimpel.
\newblock A baseline for detecting misclassified and out-of-distribution examples in neural networks.
\newblock {\em arXiv preprint arXiv:1610.02136}, 2016.

\bibitem{hendrycks2018deep}
Dan Hendrycks, Mantas Mazeika, and Thomas Dietterich.
\newblock Deep anomaly detection with outlier exposure.
\newblock {\em arXiv preprint arXiv:1812.04606}, 2018.

\bibitem{hendrycks2022pixmix}
Dan Hendrycks, Andy Zou, Mantas Mazeika, Leonard Tang, Bo~Li, Dawn Song, and Jacob Steinhardt.
\newblock Pixmix: Dreamlike pictures comprehensively improve safety measures.
\newblock In {\em Proceedings of the IEEE/CVF Conference on Computer Vision and Pattern Recognition}, pages 16783--16792, 2022.

\bibitem{inkawhich2021training}
Nathan~A Inkawhich, Eric~K Davis, Matthew~J Inkawhich, Uttam~K Majumder, and Yiran Chen.
\newblock Training sar-atr models for reliable operation in open-world environments.
\newblock {\em IEEE Journal of Selected Topics in Applied Earth Observations and Remote Sensing}, 14:3954--3966, 2021.

\bibitem{lee2020autoaudit}
Meng-Chieh Lee, Yue Zhao, Aluna Wang, Pierre~Jinghong Liang, Leman Akoglu, Vincent~S Tseng, and Christos Faloutsos.
\newblock Autoaudit: Mining accounting and time-evolving graphs.
\newblock In {\em 2020 IEEE International Conference on Big Data (Big Data)}, pages 950--956. IEEE, 2020.

\bibitem{li2025dpu}
Shawn Li, Huixian Gong, Hao Dong, Tiankai Yang, Zhengzhong Tu, and Yue Zhao.
\newblock Dpu: Dynamic prototype updating for multimodal out-of-distribution detection.
\newblock In {\em Proceedings of the IEEE/CVF Conference on Computer Vision and Pattern Recognition (CVPR)}, 2025.

\bibitem{liu2020energy}
Weitang Liu, Xiaoyun Wang, John Owens, and Yixuan Li.
\newblock Energy-based out-of-distribution detection.
\newblock {\em Advances in neural information processing systems}, 33:21464--21475, 2020.

\bibitem{marevisiting}
Longfei Ma, Yiyou Sun, Kaize Ding, Zemin Liu, and Fei Wu.
\newblock Revisiting score propagation in graph out-of-distribution detection.
\newblock In {\em The Thirty-eighth Annual Conference on Neural Information Processing Systems}.

\bibitem{mccallum2000automating}
Andrew~Kachites McCallum, Kamal Nigam, Jason Rennie, and Kristie Seymore.
\newblock Automating the construction of internet portals with machine learning.
\newblock {\em Information Retrieval}, 3:127--163, 2000.

\bibitem{mernyei2020wiki}
P{\'e}ter Mernyei and C{\u{a}}t{\u{a}}lina Cangea.
\newblock Wiki-cs: A wikipedia-based benchmark for graph neural networks.
\newblock {\em arXiv preprint arXiv:2007.02901}, 2020.

\bibitem{papadopoulos2021outlier}
Aristotelis-Angelos Papadopoulos, Mohammad~Reza Rajati, Nazim Shaikh, and Jiamian Wang.
\newblock Outlier exposure with confidence control for out-of-distribution detection.
\newblock {\em Neurocomputing}, 441:138--150, 2021.

\bibitem{qin2025metaood}
Yuehan Qin, Yichi Zhang, Yi~Nian, Xueying Ding, and Yue Zhao.
\newblock Metaood: Automatic selection of ood detection models.
\newblock In {\em International Conference on Learning Representations (ICLR)}, 2025.

\bibitem{reimers2019sentence}
N~Reimers.
\newblock Sentence-bert: Sentence embeddings using siamese bert-networks.
\newblock {\em arXiv preprint arXiv:1908.10084}, 2019.

\bibitem{sen2008collective}
Prithviraj Sen, Galileo Namata, Mustafa Bilgic, Lise Getoor, Brian Galligher, and Tina Eliassi-Rad.
\newblock Collective classification in network data.
\newblock {\em AI magazine}, 29(3):93--93, 2008.

\bibitem{song2022learning}
Yu~Song and Donglin Wang.
\newblock Learning on graphs with out-of-distribution nodes.
\newblock In {\em Proceedings of the 28th ACM SIGKDD Conference on Knowledge Discovery and Data Mining}, pages 1635--1645, 2022.

\bibitem{tao2023non}
Leitian Tao, Xuefeng Du, Xiaojin Zhu, and Yixuan Li.
\newblock Non-parametric outlier synthesis.
\newblock {\em arXiv preprint arXiv:2303.02966}, 2023.

\bibitem{vernekar2019out}
Sachin Vernekar, Ashish Gaurav, Vahdat Abdelzad, Taylor Denouden, Rick Salay, and Krzysztof Czarnecki.
\newblock Out-of-distribution detection in classifiers via generation.
\newblock {\em arXiv preprint arXiv:1910.04241}, 2019.

\bibitem{wang2025gold}
Danny Wang, Ruihong Qiu, Guangdong Bai, and Zi~Huang.
\newblock Gold: Graph out-of-distribution detection via implicit adversarial latent generation.
\newblock {\em arXiv preprint arXiv:2502.05780}, 2025.

\bibitem{wang2020further}
Ziyu Wang, Bin Dai, David Wipf, and Jun Zhu.
\newblock Further analysis of outlier detection with deep generative models.
\newblock {\em Advances in Neural Information Processing Systems}, 33:8982--8992, 2020.

\bibitem{wu2023energy}
Qitian Wu, Yiting Chen, Chenxiao Yang, and Junchi Yan.
\newblock Energy-based out-of-distribution detection for graph neural networks.
\newblock {\em arXiv preprint arXiv:2302.02914}, 2023.

\bibitem{xiao2020timme}
Zhiping Xiao, Weiping Song, Haoyan Xu, Zhicheng Ren, and Yizhou Sun.
\newblock Timme: Twitter ideology-detection via multi-task multi-relational embedding.
\newblock In {\em Proceedings of the 26th ACM SIGKDD international conference on knowledge discovery \& data mining}, pages 2258--2268, 2020.

\bibitem{xu2020cosimgnn}
Haoyan Xu, Runjian Chen, Yueyang Wang, Ziheng Duan, and Jie Feng.
\newblock Cosimgnn: towards large-scale graph similarity computation.
\newblock {\em arXiv preprint arXiv:2005.07115}, 2020.

\bibitem{xu2021graph}
Haoyan Xu, Ziheng Duan, Yueyang Wang, Jie Feng, Runjian Chen, Qianru Zhang, and Zhongbin Xu.
\newblock Graph partitioning and graph neural network based hierarchical graph matching for graph similarity computation.
\newblock {\em Neurocomputing}, 439:348--362, 2021.

\bibitem{xu2024lego}
Haoyan Xu, Kay Liu, Zhengtao Yao, Philip~S Yu, Kaize Ding, and Yue Zhao.
\newblock Lego-learn: Label-efficient graph open-set learning.
\newblock {\em Transactions on Machine Learning Research}, 2025.

\bibitem{xu2025few}
Haoyan Xu, Zhengtao Yao, Yushun Dong, Ziyi Wang, Ryan~A Rossi, Mengyuan Li, and Yue Zhao.
\newblock Few-shot graph out-of-distribution detection with llms.
\newblock {\em arXiv preprint arXiv:2503.22097}, 2025.

\bibitem{yan2023comprehensive}
Hao Yan, Chaozhuo Li, Ruosong Long, Chao Yan, Jianan Zhao, Wenwen Zhuang, Jun Yin, Peiyan Zhang, Weihao Han, Hao Sun, et~al.
\newblock A comprehensive study on text-attributed graphs: Benchmarking and rethinking.
\newblock {\em Advances in Neural Information Processing Systems}, 36:17238--17264, 2023.

\bibitem{yang2024generalized}
Jingkang Yang, Kaiyang Zhou, Yixuan Li, and Ziwei Liu.
\newblock Generalized out-of-distribution detection: A survey.
\newblock {\em International Journal of Computer Vision}, 132(12):5635--5662, 2024.

\bibitem{yang2021graphformers}
Junhan Yang, Zheng Liu, Shitao Xiao, Chaozhuo Li, Defu Lian, Sanjay Agrawal, Amit Singh, Guangzhong Sun, and Xing Xie.
\newblock Graphformers: Gnn-nested transformers for representation learning on textual graph.
\newblock {\em Advances in Neural Information Processing Systems}, 34:28798--28810, 2021.

\bibitem{zhang2023mixture}
Jingyang Zhang, Nathan Inkawhich, Randolph Linderman, Yiran Chen, and Hai Li.
\newblock Mixture outlier exposure: Towards out-of-distribution detection in fine-grained environments.
\newblock In {\em Proceedings of the IEEE/CVF Winter Conference on Applications of Computer Vision}, pages 5531--5540, 2023.

\bibitem{zhao2020uncertainty}
Xujiang Zhao, Feng Chen, Shu Hu, and Jin-Hee Cho.
\newblock Uncertainty aware semi-supervised learning on graph data.
\newblock {\em Advances in neural information processing systems}, 33:12827--12836, 2020.

\bibitem{zhu2022survey}
Yanqiao Zhu, Yuanqi Du, Yinkai Wang, Yichen Xu, Jieyu Zhang, Qiang Liu, and Shu Wu.
\newblock A survey on deep graph generation: Methods and applications.
\newblock In {\em Learning on Graphs Conference}, pages 47--1. PMLR, 2022.

\end{thebibliography}







\clearpage
\appendix

\section*{Appendix: Graph Synthetic Out-of-Distribution Exposure with
Large Language Models}

\section{ID and OOD Split}
\label{appen:OOD Split}

\begin{table}[h]
\caption{ID classes and ID ratio for different datasets.}
\label{table:split1}
\centering
\begin{tabular}{lcc}
\hline
\textbf{Dataset}       & \textbf{ID class}             & \textbf{ID ratio} \\ \hline
Cora                   & [2, 4, 5, 6]                      & 47.71\%               \\ 
Citeseer        & [0, 1, 2]                & 55.62\%               \\ 
WikiCS           & [1, 4, 5, 6]                      & 38.79\%               \\ 
Pubmed             & [0, 1] & 60.75\%               \\ \hline
\end{tabular}
\end{table}

\section{Dataset Descriptions}
\label{appen:dataset-desc}

\paragraph{Cora} The Cora dataset \cite{mccallum2000automating} contains 2,708 scientific publications categorized into seven research topics: case-based reasoning, genetic algorithms, neural networks, probabilistic methods, reinforcement learning, rule learning, and theory. Each node represents a paper, and edges correspond to citation links between papers, forming a graph with 5,429 edges.

\paragraph{CiteSeer} The CiteSeer dataset \cite{giles1998citeseer} consists of 3,186 scientific articles classified into six research domains: Agents, Machine Learning, Information Retrieval, Databases, Human-Computer Interaction, and Artificial Intelligence. Each node represents a paper, with node features extracted from the paper’s title and abstract. The graph is constructed based on citation relationships among the publications.

\paragraph{WikiCS} WikiCS \cite{mernyei2020wiki} is a Wikipedia-based graph dataset constructed for benchmarking graph neural networks. Nodes correspond to articles in computer science, divided into ten subfields serving as classification labels. Edges represent hyperlinks between articles, and node features are derived from the corresponding article texts.

\paragraph{PubMed} The PubMed dataset \cite{sen2008collective} comprises scientific articles related to diabetes research, divided into three categories: experimental studies on mechanisms and treatments, research on Type 1 Diabetes with an autoimmune focus, and Type 2 Diabetes studies emphasizing insulin resistance and management. The citation graph connects related papers, with node features derived from medical abstracts.
\section{Evaluation Metrics}
\label{appen:metrics}

We use the following metrics to evaluate in-distribution (ID) classification and out-of-distribution (OOD) detection performance, which are widely used OOD detection research \cite{dong2024multiood,li2025dpu,qin2025metaood}:

\paragraph{Accuracy (ACC)} Measures the proportion of correctly classified ID nodes:

\begin{equation}
\text{ACC} = \frac{1}{|\mathcal{D}_{\text{ID}}|} \sum_{x_i \in \mathcal{D}_{\text{ID}}} \mathbb{I}[\hat{y}_i = y_i],
\end{equation}

where $\hat{y}_i$ is the predicted class label and $y_i$ is the true class label.

\paragraph{Area Under the ROC Curve (AUROC)} Evaluates how well the OOD detector ranks OOD nodes higher than ID nodes based on their OOD scores. It is defined as:
\begin{equation}
    \text{AUROC} = \mathbb{P}\left(s_{\text{OOD}}(x_{\text{OOD}}) > s_{\text{OOD}}(x_{\text{ID}})\right),
\end{equation}
where $s_{\text{OOD}}(\cdot)$ denotes the OOD score function.

\paragraph{Area Under the Precision-Recall Curve (AUPR)} Measures the area under the curve defined by precision and recall:
\begin{align}
    \text{Precision} &= \frac{\text{TP}}{\text{TP} + \text{FP}}, \\
    \text{Recall} &= \frac{\text{TP}}{\text{TP} + \text{FN}}, \\
    \text{AUPR} &= \int_0^1 \text{Precision}(r) \, dr,
\end{align}
where OOD nodes are treated as the positive class, and $r$ denotes recall.

\paragraph{False Positive Rate at 95\% True Positive Rate (FPR@95)} Indicates the fraction of ID samples incorrectly predicted as OOD when the true positive rate (TPR) on OOD samples is 95\%:
\begin{equation}
    \text{FPR@95} = \frac{\text{FP}_{\text{ID}}}{\text{FP}_{\text{ID}} + \text{TN}_{\text{ID}}} \Bigg|_{\text{TPR}=0.95},
\end{equation}
where FP and TN are false positives and true negatives on ID data, respectively.

\section{Standard Deviation Results}
\label{appen::STD}

\begin{table*}[h]
    \centering
    \caption{Standard deviation results for various models across ID classification and OOD detection metrics. All values are in percentages.}
    \label{tab:std results}
    \renewcommand{\arraystretch}{1.5} 
    \setlength{\tabcolsep}{1pt}      
    \resizebox{\textwidth}{!}{
    \begin{tabular}{llcccccccccccccccc}
        \toprule
        \multirow{2}{*}{\textbf{Model}} & \multirow{2}{*}{\textbf{Methods}} 
            & \multicolumn{4}{c}{\textbf{Cora}} 
            & \multicolumn{4}{c}{\textbf{Citeseer}} 
            & \multicolumn{4}{c}{\textbf{Pubmed}} 
            & \multicolumn{4}{c}{\textbf{Wiki-CS}} \\
        \cmidrule(lr){3-6}\cmidrule(lr){7-10}\cmidrule(lr){11-14}\cmidrule(lr){15-18}
        & & ACC & AUROC  & AUPR & FPR95 
          & ACC & AUROC & AUPR & FPR95 
          & ACC  & AUROC  & AUPR & FPR95
          & ACC & AUROC & AUPR & FPR95 \\
        \midrule
        \multirow{5}{*}{\shortstack{No OOD\\Exposure}} 
        & MSP 
            & 2.46 & 1.90 & 1.47 & 7.68
            & 1.39 & 2.85 & 3.04 & 5.38 
            & 3.06 & 2.12 & 3.02 & 2.37
            & 2.73 & 3.65 & 4.19 &6.74 \\
        & Entropy 
            & 2.18 & 2.78 & 1.87 & 8.91 
            & 1.39 & 2.59 & 2.80 & 6.31
            & 3.06 & 2.12 & 3.02 & 2.37
            & 2.71 & 3.33 & 3.71 & 7.10 \\
        & Energy 
            & 2.17 & 2.18 & 1.58 & 9.95
            & 1.61 & 2.54 & 2.27 & 5.68
            & 1.13 & 8.70 & 9.23 & 2.82 
            & 2.67 & 5.70 & 4.92 & 13.74 \\
        & GNNSafe 
            & 2.18 & 1.74 & 2.12 & 10.52
            & 1.61 & 2.58 & 2.36 & 5.91
            & 1.23 & 14.34 & 13.08 & 8.26
            & 2.63 & 0.80 & 0.90 & 16.60 \\
        & GRASP 
            & 2.18 & 1.88 & 1.34 & 13.29 
            & 1.62 & 2.88 & 2.82 & 6.92 
            & 1.32 & 12.34 & 11.76 & 5.70 
            & 2.92 & 2.58 & 3.18 & 12.96 \\
        \midrule
        \multirow{2}{*}{Ours}
        & \methodid 
            & 2.53 & 0.88 & 1.42 & 2.24
            & 1.65 & 2.01 & 1.85 & 7.66
            & 2.03 & 2.48 & 3.02 & 7.69 
            & 2.20 & 0.90 & 1.26 & 11.86 \\
        & \methodgen 
            & 2.18 & 1.67 & 2.12 & 7.03
            & 1.49 & 1.31 & 1.58 & 7.12
            & 2.69 & 6.86 & 5.62 & 7.79
            & 2.66 & 1.22 & 1.09 & 1.62 \\
        \midrule
        \multirow{2}{*}{\shortstack{Real OOD\\Exposure}} 
        & OE & 1.86 & 1.19 & 1.03 & 6.39  & 1.60 & 2.57 & 2.24 & 11.08 & 1.88 & 5.17 & 5.27 & 14.78 & 1.40 & 1.64 & 1.71 & 9.08\\
        & GNNSafe++ 
            & 2.57 & 0.93 & 1.46 & 3.84
            & 1.47 & 2.08 & 2.22 & 8.69
            & 1.26 & 3.83 & 3.45 & 1.13
            & 2.41 & 1.04 & 1.32 & 12.27\\
        \bottomrule
    \end{tabular}
    }
\end{table*}

\newpage
\section{OOD Detection Methods Comparison}
\label{appen:methods comparison}
\begin{table}[ht]
\centering
\scriptsize
\setlength{\tabcolsep}{4pt}
\renewcommand{\arraystretch}{1.5}
\begin{tabularx}{\textwidth}{|l|X|c|X|c|l|}
\hline
\textbf{Method} & \textbf{Core Technique} & \textbf{Uses LLM?} & \textbf{LLM Role (if any)} & \textbf{Requires Real OOD Data?} & \textbf{Data Type} \\ \hline
GNNSafe \cite{wu2023energy} & Post-hoc Energy Scoring & No & N/A & No & Graph \\ \hline
GRASP \cite{marevisiting} & Post-hoc Score Propagation & No & N/A & No & Graph \\ \hline
GNNSafe++ \cite{wu2023energy} & OOD Exposure (Real Data) + Energy Regularization & No & N/A & Yes & Graph \\ \hline
VOS \cite{du2022vos} & Generative Pseudo-OOD (Representations) & No & N/A & No & Image \\ \hline
GOLD \cite{wang2025gold} & Implicit Adversarial Pseudo-OOD Latent Generation & No & N/A & No & Graph \\ \hline
Synthetic \cite{abbas2025out} & LLM-Generated Pseudo-OOD Proxies & Yes & Pseudo-OOD Text Generation & No & Text \\ \hline
EOE \cite{cao2024envisioning} & LLM-Envisioned Outlier Exposure & Yes & Outlier Concept Generation & No & Image \\ \hline
\methodid & OOD Exposure (LLM-Identified Pseudo-OOD Nodes) & Yes & Pseudo-OOD Node Identification & No & Text-Attributed Graph \\ \hline
\methodgen & OOD Exposure (LLM-Generated Pseudo-OOD Nodes) & Yes & Pseudo-OOD Node Generation & No & Text-Attributed Graph \\ \hline
\end{tabularx}
\caption{Comparison of various OOD detection methods.}
\label{tab:ood_methods}
\end{table}

\end{document}